\documentclass[runningheads]{llncs}

\usepackage{eccv}

\usepackage{eccvabbrv}

\usepackage{graphicx}
\usepackage{booktabs}
\usepackage{multicol}
\usepackage{multirow}
\usepackage{tabularx}
\usepackage{wrapfig}
\usepackage{amsfonts}
\usepackage{amsmath}
\usepackage{amssymb}
\usepackage{bm}
\usepackage[accsupp]{axessibility}  

\usepackage{hyperref}

\usepackage{orcidlink}

\begin{document}

\title{Adapting Large Language Model for Cross-Subject Semantic Decoding from Video-Stimulated fMRI}
\titlerunning{Abbreviated paper title}

\author{Ruizhe Zheng\inst{1} \and
Lichao Sun\inst{2}
}

\authorrunning{R. Zheng, L. Sun}

\institute{Fudan University, 220 Handan Road, Shanghai 200433 China \and Lehigh University, 27 Memorial Drive West Bethlehem, PA 18015 USA \\ 
\email{rzzheng23@m.fudan.edu.cn}, \email{james.lichao.sun@gmail.com}\\
}
\maketitle

\begin{abstract}
    Decoding visual-semantic information from brain signals, such as functional MRI (fMRI), across different subjects poses significant challenges, including low signal-to-noise ratio, limited data availability, and cross-subject variability. Recent advancements in large language models (LLMs) show remarkable effectiveness in processing multimodal information. In this study, we introduce an LLM-based approach for reconstructing visual-semantic information from fMRI signals elicited by video stimuli. Specifically, we employ fine-tuning techniques on an fMRI encoder equipped with adaptors to transform brain responses into latent representations aligned with the video stimuli. Subsequently, these representations are mapped to textual modality by LLM. In particular, we integrate self-supervised domain adaptation methods to enhance the alignment between visual-semantic information and brain responses. Our proposed method achieves good results using various quantitative semantic metrics, while yielding similarity with ground-truth information. 
\end{abstract}

\section{Introduction}
\label{sec:intro}

Advancements in semantic brain decoding, which aims at reconstructing the semantic information implicitly contained in various external stimuli such as image and video signals from brain activity patterns, have showcased the remarkable potential of decoding information that offers a pathway to mind-reading technologies that can have important clinical and scientific applications\cite{horikawa_generic_2017,ozcelik_reconstruction_2022}. However, such endeavors carry many challenges. Noninvasive imaging techniques such as functional MRI (fMRI) have lower temporal or spatial resolution and varies a lot across different individuals due to unique anatomical and functional attributes. The rarity of data is also an important concern as the brain decoders are often insufficiently trained and may have problem generalizing on newly encountered subjects with different condition and content of stimuli. Despite breakthroughs in artificial intelligence for brain decoding, these limitations raises questions regarding generalizability and effectiveness of cross-subject reconstruction of visual-semantic information.

In recent years, large language models (LLMs) have continuously pushed the upper limit of natural language understanding with ever increasing parameter sizes and pre-training data scales\cite{video-chatgpt}\cite{videochat}. In particular, LLMs have demonstrated remarkable multimodal information processing and have achieved great success in generating visual-semantic contents. In terms of visual understanding, by conditioning the model with one or more modalities or instructions, LLMs can achieve strong few-shot or zero-shot performance on vision-language tasks. In a lot of scenarios, the LLM-generated texts have high quality and fidelity and cannot easily be distinguished from genuine human texts. Adapting LLMs for visual understanding is computationally intensive, resulting in considerable memory consumption. Therefore, many researchers have attempt to apply frozen pre-trained language decoders and vision encoders in accommodating visual input. Image- or video-text cross-modal learning has achieved remarkable performance in many downstream tasks by using various computationally efficient strategies that optimizes a small proportion of parameters or additionally equipped adaptor modules. For instance, BLIP-2\cite{blip} uses a small Transformer-based adaptor during vision-language alignment training and instruction tuning. Video-LLaMA\cite{lin2023videollava} applies similar approach in video-language representation. The advantage of this solution is that it takes advantage of existing models and requires only parameter-efficient tuning rather than full finetuning of LLMs and visual encoders.

Inspired by these observations, we propose a novel multimodal finetuning framework that fully leverages frozen brain and visual encoders, coupled with an instruction-tuned video-language foundation model, to decode linguistic representations from brain signals recorded in subjects who receive dynamic visual stimuli. Given the large size of raw fMRI data and the intrinsic spatio-temporal dynamics, we design a three-dimensional Convolutional Neural Network (CNN) tokenizer to transform raw fMRI data into tokens which will be further encoded by a brain encoder pretrained on Human Connectome Project (HCP) datasets. In the first stage, it jointly optimizes low-rank adaptors, which are attached to fMRI encoder and video Q-former, along with projection adaptors that connect the intermediate embedding of fMRI with LLM, and the spatio-temporal tokenizer, to learn visual-linguistic patterns from raw neural data by contrastive learning. In the second stage, as we lack groundtruth linguistic representation of video contents, we sample texts from Video-LLaMA, one of the state-of-the-art multimodal LLMs for video understanding. Then, video query tokens concatenated with pertinent video-related questions are processed by LLaMA. The generated answers can be used to construct paired fMRI-text data for supervised instruction finetuning. During inference, LLaMA will receive only text prompts and fMRI tokens for comprehension of the visual-semantic brain activities. Moreover, we employ a self-supervised domain adaptation approach to learn resilient, discriminative feature embeddings across individuals while preventing the inadvertent leakage of visual information. Importantly, the entire training procedure remains agnostic to both the stimuli and their corresponding labels within the validation set while reducing the subject-wise domain discrepancies. 

We summarize the contributions of the work as follows. First, an end-to-end LLM-centric pipeline is established to replace traditional multimodal neural networks. Despite no groundtruth semantic information is available, we manage to use LLM for automatic annotation to create aligned fMRI-video-text triads. Second, we investigate on video rather than image modality, which further increases the difficulty because both spatial and temporal information is required for holistic visual understanding. Third, our method demonstrates good generalizability on distinct individuals and stimuli, which is of pivotal importance in neuroscientific research and applications.

\section{Related Work}
Many previous work focus on direct reconstruction of stimulating signals by mapping latent representation of fMRI and visual signals from one to the other. For instance, \cite{horikawa_generic_2017} uses linear regression regularized by sparsity constraints on preprocessed fMRI data to predict features extracted from low-level neural representation by pretrained CNN for images, which does not involve more subtle alignment of distinct modalities. \cite{shen_end--end_nodate} uses fMRI data and stimulus images to create an end-to-end reconstruction model involing training a generative adversarial network (GAN). \cite{ozcelik_reconstruction_2022} uses conditioned GAN to reconstruct images that are consistent with groundtruth in terms of semantic meanings. \cite{takagi2022} apply diffusion model guided by semantic information of image content to reconstruct image from fMRI. \cite{chen2023cinematic} manages to reconstruct video from fMRI by using multimodal alignment to extract semantically rich representations as guidance for diffusion-based decoding. High-level or semantic information reconstruction, which is the main task investigated in the paper, involves more complicated techniques which requires more discriminative representation of stimulating signals. \cite{ferrante2023brain} have investigated captioning of image data by utilizing a combination of a pre-trained visual encoder and language decoder for semantic reconstruction. \cite{tang2023semantic} use similar approach to reconstruct intelligible word sequences that recover the semantic representation of speech and video stimuli in human brains.
\section{Methods}
\subsection{Architecture}
\noindent\textbf{Foundation Models.}~~
We follow the Video-LLaMA architecture \cite{videollama}. A Query-Former is an encoder-only transformer with 32 learned query tokens as input: it contextualizes the query tokens – via the cross-attention mechanism – with the representations of the image patches encoded by a large (frozen) Vision Transformer (ViT) The visual tokens that are the output of the Q-Former are then projected into the LLM embedding space with a single linear projection. For fMRI data, we use Sparse-Coded Masked Brain Modeling (SC-MBM) model \cite{chen2023cinematic}, a fMRI encoder developed on large amount of data downloaded from Human Connectome Projects (HCP), which contains high-quality fMRI recorded under resting and task-evoked paradigm. The encoder is pretrained through a masked autoencoder strategy, such that the model will acquire strong contextual abstraction of the temporal and spatial associations by enforcing it to restore the masked voxels.

\noindent\textbf{Spatio-Temporal Convolutional Tokenizer.}~~
To make fully usage of the fMRI data, we decide to not restrict our analysis to the visual cortex, since other cortical regions involving in visual and semantic information flow may also be beneficial for decoding. However, each block of fMRI data contains more than 1,000,00 voxels in total in our research. This renders the tackling of spatio-temporal sequence by voxel-based SC-MBM very challenging because the self-attention mechanism demands quadratic memory consumption. Therefore, we are motivated to design a three-dimensional convolutional tokenizer that will transform the huge amount of voxels into super-voxel sequence representation suitable for further processing by SC-MBM.

\noindent\textbf{Design of Adaptor.}~~ 
Finetuning large pretrained models for downstream adaptations has become a standard technical pipeline. Recent research \cite{hu2021lora, yin20231, yuan2024fullloraat} have highlighted the feasibility of tuning models by freezing the pre-trained parameters and introducing usually structured small amount of new parameters to the original architecture. As shown in Figure 2, we design a nonlinear low-rank adaptor of parameter-efficient repurposing of the multimodal combination of models for our goal. By inserting these adaptors into the query projection layers in self-attention and the multilayer perceptron modules of the ViT-based fMRI encoder and the BERT Transformer-based Q-Former.

\begin{figure*}
\begin{center}
\centerline{\includegraphics[width=0.9\textwidth]{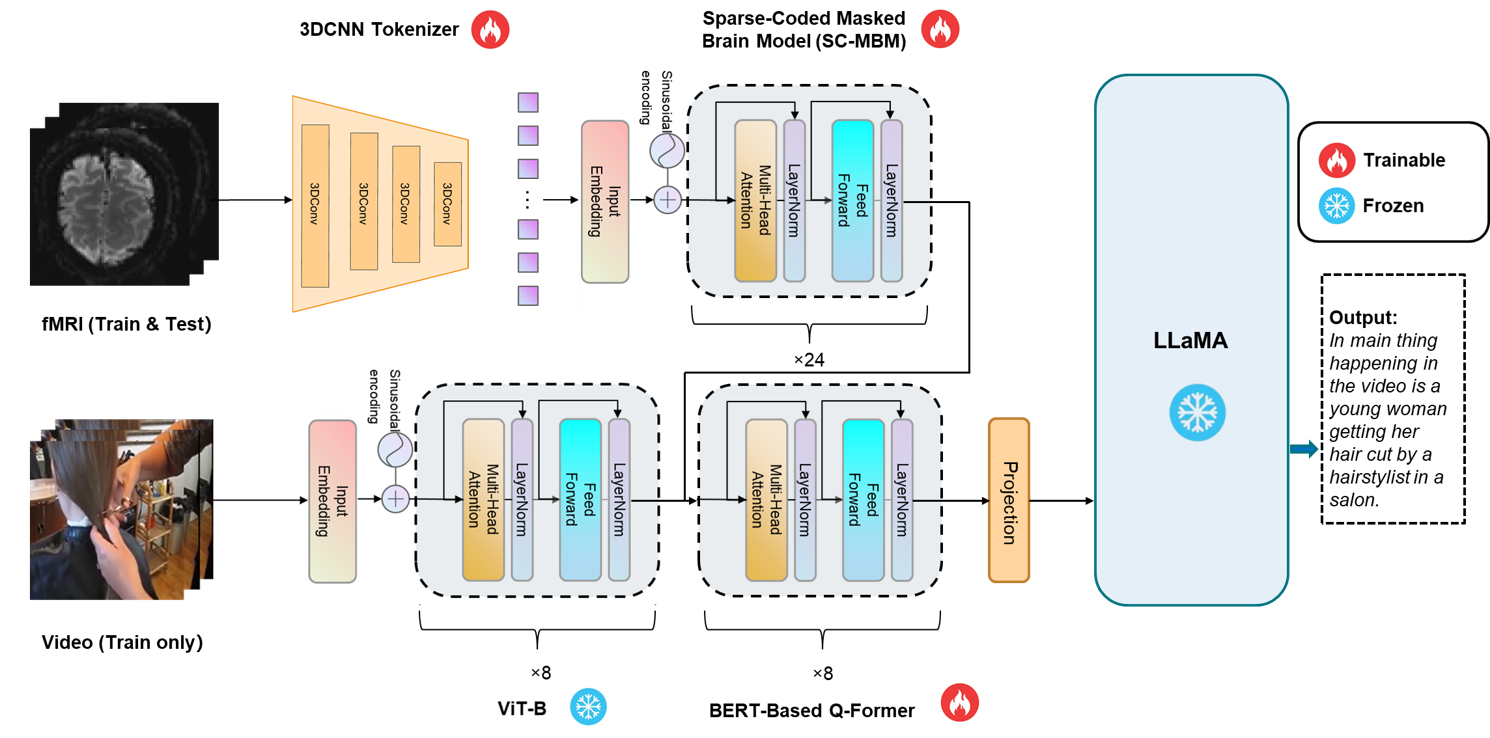}}
\caption{The overall framework of our approach for brain  visual-semantic reconstruction. The fMRI is encoded by a 3DCNN tokenizer and SC-MBM. The video is encoded by ViT. The parameters of SC-MBM, ViT and Q-Former are all frozen, but SC-MBM and Q-Former is inserted with the nonlinear adaptor module. During training, it learns cross-subject semantically informed fMRI latent representation by cross-modal alignment and domain adaptation, and the quality of decoding is improved by minimizing the difference between video- and fMRI-based video understanding by the instruction-tuned LLM.
}
\label{fig:vis-bold}
\end{center}
\end{figure*}
\vskip -0.25in
\subsection{Training Procedure} Our approach adopts a two-stage training paradigm for training the model: cross-modal alignment (Stage I) and supervised instruction fine-tuning (Stage II). Let $X$ be the video embedding of $L$ tokens calculated by Q-Former, $Z$ be the fMRI embedding calculated by fMRI encoder equipped with adaptors, we formulate training objectives as follows.

\noindent\textbf{Stage I.}~~
In this stage, we aim at learng a cross-modal alignment between the embedding spaces of fMRI and video from corpora of their respective modalities. We aim to distinguish the right fMRI patterns out of a batch of data, each contains different neural visual and semantic representations. To do so, we adopt the CLIP loss proposed in \cite{radford2021learning}. Specifically, we conduct cross-modal alignment by drawing the paired video and fMRI embeddings extracted by SC-MBM together while pushing unpaired away in the latent space.

To do so, we train f$_\mathrm{\theta}$ using the CLIP loss 
on batches of size B with exactly one positive example:
$\begin{pmatrix}1\end{pmatrix}$ 
$$
\mathcal{L}_{CLIP}(\theta)=-\frac{1}{B}\sum_{i=1}^B\left(\log\frac{\exp(s(\hat{\boldsymbol{z}}_i,\boldsymbol{z}_i)/\tau)}{\sum_{j=1}^B\exp(s(\hat{\boldsymbol{z}}_i,\boldsymbol{z}_j)/\tau)}+\log\frac{\exp(s(\hat{\boldsymbol{z}}_i,\boldsymbol{z}_i)/\tau)}{\sum_{k=1}^B\exp(s(\hat{\boldsymbol{z}}_k,\boldsymbol{z}_i)/\tau)}\right)
$$ where $s$ is the cosine similarity, $z_i$ and $\hat{z}_i=\mathbf{f}_\theta(X_i)$ are the latent video representation and the corresponding fMRI-based prediction, respectively, and $\tau$ is a learned temperature parameter, which is set as 0.05 during training.

Next, as we also want the LLM to process fMRI tokens so as to extract immanent visual-language information, we also train adaptors to directly map the fMRI to visual-language embeddings such that they can be understood by the frozen LLM to reconstruct individual visual-language cognition of video stimuli. This is achieved using weighted $L_2$ and $L_1$ reconstruction losses:
\begin{equation}
\mathcal{L}_{L_2-L_1}(\theta)=\frac{1}{NL}\sum_{n=1}^{N}\sum_{l=1}^{L}(1-\alpha)\lVert z_{n}^{(l)}-x_{n}^{(l)}\rVert_{2}^{2} + \alpha\lVert z_{n}^{(l)}-x_{n}^{(l)}\rVert_{1}
\end{equation}
Finally, we combine the CLIP and reconstruction losses using a convex combination with tuned weight to train models that benefit from both training objectives:
\begin{equation}
\mathcal{L}_{Total}=\beta\mathcal{L}_{CLIP}+(1-\beta)\mathcal{L}_{L_2-L_1}
\end{equation}
\noindent\textbf{Stage II.}~~
As there is no groundtruth for video-language understanding in our experimental setting, we adopt a bootstrapping approach for training the model to reconstruct semantic information from video stimuli-induced fMRI activities. For a given batch of $N$ fMRI records, we assign each of them a randomly selected instruction from a candidate instruction list, and generate surrogate groundtruth data from ViT-embedded video tokens and the instructions. These surrogate texts are used for supervised instruction tuning that will allow the model to learn more intricate semantic information. Then, we freeze all parameters of LLM, Qformer and fMRI encoder except for an adaptor that bridges the encoders and the LLM. The adaptor is trained with cross-entropy loss
\begin{equation}
\mathcal{L}_{\text{CE}} = -\frac{1}{B} \sum_{j=1}^{B} \sum_{t=1}^{T} \log p(y_{j,t} | y_{j<t}, \theta),
\end{equation}
where $y_{j,t}$ denotes the true token at position $t$ in the $j$-th sequence in the batch, $y_{j<t}$ represents the tokens preceding $y_{j,t}$.

We introduce an additional classifier head to the original framework to conduct domain adaptation. Specifically, a neighborhood clustering-based approach \cite{saito2020universal} applied in order to learn better fMRI representation from a proportion of target domain data. Let $\mathbf{g}$ be a trainable linear projection layer. Its weight $[\mathbf{w}_1, \mathbf{w}_2, \dots, \mathbf{w}_C]$ is conceived as $C$ video classes contained in the training data. $y$ is the output of projection after SoftMax activation, or the predicted categorical distribution. According to the approach proposed in \cite{saito2020universal}, a memory pool that stores $N$ target domain feature vectors is to be trained and concatenated with the weight vectors of $\mathbf{g}$. Then the similarity of $i$-th ($i \neq j)$ target domain feature $f_i$ to the memory features $F_i$ is formulated as $p_{i,j} = \frac{\exp ({\bm{F_j}}^\top \bm{f}_i / \tau)}{Z_i},$, where$Z_i = {\sum_{j=1, j \neq i}^{N + C} \text{exp}({\bm{F_j}}^\top \bm{f}_i / \tau) }.$ The scale parameter $\tau$ is set as 0.5 in our experiments. Thus, the proportion of fMRI data from validation subjects are used to calculate a domain alignment loss, which is called neighborhood clustering (NC) loss in \cite{saito2020universal}, so as to minimize the discrepancies inevitably encountered in the proprotion of fMRI data that will be used for test:
\begin{equation}
\mathcal{L}_{\text{NC}} = - \frac{1}{B}\sum_{i=1}^{B} \sum_{j=1, j \neq i}^{N + K} p_{i,j}\log(p_{i,j}),
\label{eq:predictions}
\end{equation}
where $B$ is batch size.

Because we assume that we have no prior knowledge of the categorical information of test data, we also apply the entropy separation (ES) loss proposed in \cite{saito2020universal} that will be optimized to make the entropy of target sample semantic classes larger and the source samples smaller. Let $\bm{p}_i$ be the predicted class probability vector, $m$ and $rho$ are hyperparameters, the loss is formulated as
\begin{equation}
\mathcal{L}_{\text{ES}} = \frac{1}{B}\sum_{i=1}^B \mathcal{L}_{\text{ES}}  (\bm{p}_{i}),\ \  \mathcal{L}_{\text{es}}  (\bm{p}_{i}) = \begin{cases}
    -|H(\bm{p}_{i}) - \rho| & (|H(\bm{p}_{i}) - \rho| > m), \\
    0 & otherwise.
  \end{cases}
  \label{eq:ent_sepa}
\end{equation}
To be clear, no target domain videos or annotation labels are used during training. The final evaluation is conducted on the rest proprotion of test subjects. Therefore, we boost better fMRI learning by utilizing the strong generation capabilities of LLM as well as domain adapation. The training objective is set as
\begin{equation}
    \mathcal{L}_{Stage II}=\lambda\mathcal{L}_{Cross Entropy}+(1-\lambda)\mathcal{L}_{Domain Adaptation}
\end{equation}

\section{Experiments}
\subsection{Datasets}
Throughout this work, we use two openly available fMRI-video datasets. They were collected and published with all participants read and signed an informed consent form approved by the respective ethics committee. No identifiable subject information is contained in the data. 

\begin{wrapfigure}{r}{7cm}
\includegraphics[width=0.56\textwidth]{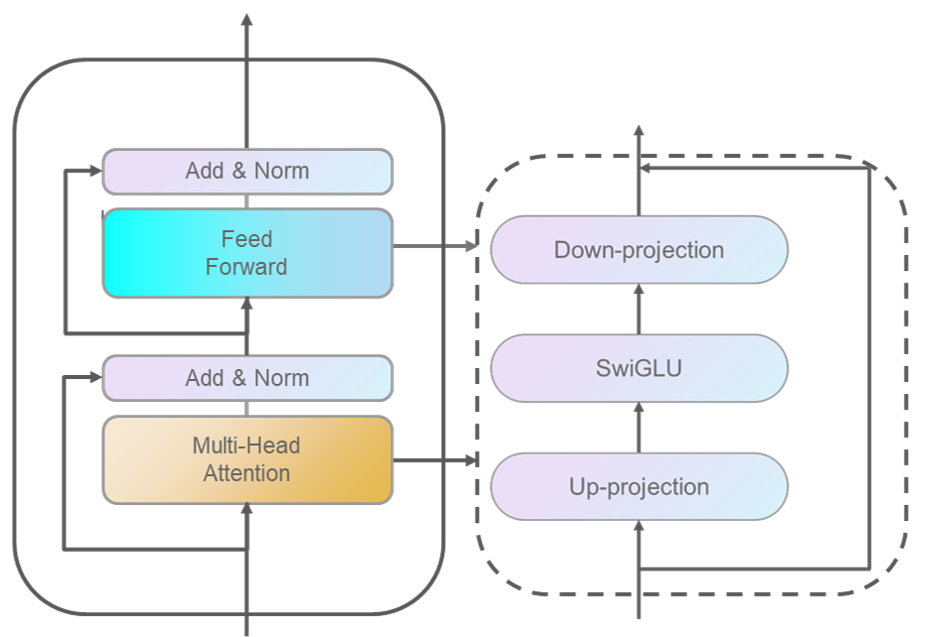}
\caption{The nonlinear adaptor used for finetuning.}
\label{fig:vis-bold}
\end{wrapfigure}
\noindent\textbf{Large-Scale fMRI Human Action Recognition Dataset}. This dataset is described in \cite{zhou2023large}. It is a large-scale fMRI dataset for human action recognition consisting of fMRI responses to 21,600 video clips from 30 participants. The video clips encompass 180 human action categories and offer a comprehensive coverage of complex activities in daily life. 26 subjects are assigned as training data and 4 subjects are used for validation.

\noindent\textbf{Urgen Natural Human Action Dataset}. This dataset is described in \cite{urgen2022large}. It is a fMRI dataset recorded on 4 subjects under visual stimuli randomly sampled from a large video set consisting of 100 different natural actions. 2 subjects are assigned as training data and 2 subjects are used for validation. 

\subsection{Results}  
As shown in Table 1 and 2, our approach achieves effective results among different individuals, resulting in the average BERTScore and SacredBLEU-1 across all validation individuals of 53.27\% and 33.91\% on Large-Scale fMRI Human Action Recognition Dataset. Further validation of our method on the Urgen Natural Human Action Dataset achieve results of 66.10\% and 53.59\%, respectively. The outcomes suggest the importance of using strong LLM as semantic decoder and the effectiveness of both appropriate multimodal alignment training and domain adaption in finetuning the fMRI-encoder so that it can actually retrieve consistent video-evoked brain responses. 

\begin{table}
\scriptsize{
\begin{tabular*}{\hsize}{@{}@{\extracolsep{\fill}}|c|c|c|c|c|c|c| @{}}
    \hline
    \hline
    \multirow{2}{*}{\textbf{Subject}} & 
    \multirow{1}{*}{\textbf{BERTScore (\%)}} &
    \multicolumn{2}{c}{\textbf{SacredBLEU (\%)}}&
    \multicolumn{3}{c}{\textbf{Rouge-L (\%)}}
    \\
     &&\textbf{SacredBLEU-1} &\textbf{SacredBLEU-2} & \textbf{F} &\textbf{P} &\textbf{R}\\
    \hline
    
    Subject-1 & 54.48 & 32.92 & 22.35 &33.97&35.44&40.02
    \\
    \hline
    Subject-2	&51.77 & 34.87&	23.27&34.22 &	37.11&35.82
\\
    \hline
    Subject-3	&52.01 &33.50 &	22.71 &	34.35 &	36.61 & 38.36
\\
    \hline
    Subject-4	&54.81 &	34.34 &	23.44 &	34.30 	&33.86 &41.48 \\
    \hline
    Total & 66.10 &53.59 &43.98 & 53.22& 54.47& 54.55\\
    \bottomrule
\end{tabular*}
}
  \caption{Results of BERTScore, SacredBLEU, and Rouge-L metrics for fMRI semantic reconstruction on \textit{Large-Scale fMRI Human Action Recognition Dataset}. }
\end{table}
\begin{table}
\scriptsize{
\begin{tabular*}{\hsize}{@{}@{\extracolsep{\fill}}|c|c|c|c|c|c|c| @{}}
    \hline
    \hline
    \multirow{2}{*}{\textbf{Subject}} & 
    \multirow{1}{*}{\textbf{BERTScore (\%)}} &
    \multicolumn{2}{c}{\textbf{SacredBLEU (\%)}}&
    \multicolumn{3}{c}{\textbf{Rouge-L (\%)}}
    \\
     &&\textbf{SacredBLEU-1} &\textbf{SacredBLEU-2} & \textbf{F} &\textbf{P} &\textbf{R}\\
    \hline
    Subject-1 & 67.14 &  54.19& 45.00 &54.42&56.49&54.99
    \\
    \hline
    Subject-2	&65.06 & 52.98&	42.95& 52.02 &55.96&54.11\\
    \hline
    Total & 66.10&53.59&43.98&53.22&54.47&54.55\\
    \bottomrule
  \end{tabular*}}
  \caption{Results of BERTScore, SacredBLEU, and Rouge-L metrics for fMRI semantic reconstruction on \textit{Urgen Natural Human Action Dataset}.}
\end{table}
We showcase some reconstructed semantic information from fMRI with the groundtruth video understanding prompted by several questions. The current approach demonstrates strong temporal understanding ability, which is crucial for accurate visual-semantic decoding as the actions and scenes in the video continually changes by time order. 
\begin{itemize}
    \item  
    \textbf{Question}: What is the main thing happening in the video?
    \item
    \textbf{Video2Text}:\textit{In the video, a young man is playing squash in a court. He is wearing a white shirt and is holding a tennis racket. He is hitting the ball with the racket and running around the court. }\\
    \textbf{fMRI2Text}:\textit{Based on the visual content, the main thing happening in the video is a young man playing squash in a gym. He is wearing a black shirt and is playing with a racket. He is hitting the ball against the wall of the gym, and the ball is bouncing back and forth between him and the wall.}\\
    \item
    \textbf{Video2Text}:\textit{ In the video, we see a young woman getting her hair cut by a hairdresser in a salon. The hairdresser is cutting her hair with scissors, and the woman is sitting in a chair. }\\
    \textbf{fMRI2Text}:\textit{ In main thing happening in the video is a young woman getting her hair cut by a hairstylist in a salon.} \\
\end{itemize}
\section{Conclusion}
In this study, we highlight the strong capacity of Large Language Models (LLMs) to reconstruct visual-semantic information from fMRI brain responses. Our findings demonstrate that the proposed methodology can produce summaries of video content in a manner that is independent of both the subject and the stimuli. This research presents a novel approach to semantic decoding, which holds promise for applications in brain-machine interfaces and the investigation of human brain response characteristics to alterations in external stimuli. Drawing on these foundations, we could further our comprehension of how the human brain processes visual and linguistic information, ultimately enhancing generative AI and its associated applications.

%
%
\bibliographystyle{splncs04}
\bibliography{main}

\begin{thebibliography}{10}
\providecommand{\url}[1]{\texttt{#1}}
\providecommand{\urlprefix}{URL }
\providecommand{\doi}[1]{https://doi.org/#1}

\bibitem{chen2023cinematic}
Chen, Z., Qing, J., Zhou, J.H.: Cinematic mindscapes: High-quality video reconstruction from brain activity (2023)

\bibitem{ferrante2023brain}
Ferrante, M., Ozcelik, F., Boccato, T., VanRullen, R., Toschi, N.: Brain captioning: Decoding human brain activity into images and text (2023)

\bibitem{horikawa_generic_2017}
Horikawa, T., Kamitani, Y.: Generic decoding of seen and imagined objects using hierarchical visual features. Nature Communications  \textbf{8}(1),  15037 (Aug 2017). \doi{10.1038/ncomms15037}, \url{http://www.nature.com/articles/ncomms15037}

\bibitem{hu2021lora}
Hu, E.J., Shen, Y., Wallis, P., Allen-Zhu, Z., Li, Y., Wang, S., Wang, L., Chen, W.: Lora: Low-rank adaptation of large language models. arXiv preprint arXiv:2106.09685  (2021)

\bibitem{blip}
Li, J., Li, D., Xiong, C., Hoi, S.: Blip: Bootstrapping language-image pre-training for unified vision-language understanding and generation. In: ICML. pp. 12888--12900. PMLR (2022)

\bibitem{videochat}
Li, K., He, Y., Wang, Y., Li, Y., Wang, W., Luo, P., Wang, Y., Wang, L., Qiao, Y.: Videochat: Chat-centric video understanding. arXiv preprint arXiv:2305.06355  (2023)

\bibitem{lin2023videollava}
Lin, B., Ye, Y., Zhu, B., Cui, J., Ning, M., Jin, P., Yuan, L.: Video-llava: Learning united visual representation by alignment before projection (2023)

\bibitem{video-chatgpt}
Maaz, M., Rasheed, H., Khan, S., Khan, F.S.: Video-chatgpt: Towards detailed video understanding via large vision and language models. arXiv preprint arXiv:2306.05424  (2023)

\bibitem{ozcelik_reconstruction_2022}
Ozcelik, F., Choksi, B., Mozafari, M., Reddy, L., VanRullen, R.: Reconstruction of {Perceived} {Images} from {fMRI} {Patterns} and {Semantic} {Brain} {Exploration} using {Instance}-{Conditioned} {GANs} (Feb 2022), \url{http://arxiv.org/abs/2202.12692}, arXiv:2202.12692 [cs, eess, q-bio]

\bibitem{radford2021learning}
Radford, A., Kim, J.W., Hallacy, C., Ramesh, A., Goh, G., Agarwal, S., Sastry, G., Askell, A., Mishkin, P., Clark, J., et~al.: Learning transferable visual models from natural language supervision. In: International conference on machine learning. pp. 8748--8763. PMLR (2021)

\bibitem{saito2020universal}
Saito, K., Kim, D., Sclaroff, S., Saenko, K.: Universal domain adaptation through self supervision. Advances in neural information processing systems  \textbf{33},  16282--16292 (2020)

\bibitem{shen_end--end_nodate}
Shen, G., Dwivedi, K., Majima, K., Horikawa, T., Kamitani, Y.: End-to-end deep image reconstruction from human brain activity. Front. Comput. Neurosci.  \textbf{13}, ~21 (Apr 2019)

\bibitem{takagi2022}
Takagi, Y., Nishimoto, S.: High-resolution image reconstruction with latent diffusion models from human brain activity. bioRxiv  (2023). \doi{10.1101/2022.11.18.517004}, \url{https://www.biorxiv.org/content/early/2023/03/11/2022.11.18.517004}

\bibitem{tang2023semantic}
Tang, J., LeBel, A., Jain, S., Huth, A.G.: Semantic reconstruction of continuous language from non-invasive brain recordings. Nature Neuroscience pp.~1--9 (2023)

\bibitem{urgen2022large}
Urgen, B.A., Nizamo{\u{g}}lu, H., Ero{\u{g}}lu, A., Orban, G.A.: A large video set of natural human actions for visual and cognitive neuroscience studies and its validation with fmri. Brain Sciences  \textbf{13}(1), ~61 (2022)

\bibitem{yin20231}
Yin, D., Yang, Y., Wang, Z., Yu, H., Wei, K., Sun, X.: 1\% vs 100\%: Parameter-efficient low rank adapter for dense predictions. In: Proceedings of the IEEE/CVF Conference on Computer Vision and Pattern Recognition. pp. 20116--20126 (2023)

\bibitem{yuan2024fullloraat}
Yuan, Z., Zhang, J., Shan, S.: Fulllora-at: Efficiently boosting the robustness of pretrained vision transformers (2024)

\bibitem{videollama}
Zhang, H., Li, X., Bing, L.: Video-llama: An instruction-tuned audio-visual language model for video understanding. arXiv preprint arXiv:2306.02858  (2023)

\bibitem{zhou2023large}
Zhou, M., Gong, Z., Dai, Y., Wen, Y., Liu, Y., Zhen, Z.: A large-scale fmri dataset for human action recognition. Scientific Data  \textbf{10}(1), ~415 (2023)

\end{thebibliography}
\end{document}